\pdfoutput=1

\documentclass[11pt]{article}

\usepackage[final]{acl}

\usepackage{times}
\usepackage{latexsym}

\usepackage[T1]{fontenc}

\usepackage[utf8]{inputenc}

\usepackage{microtype}

\usepackage{inconsolata}

\usepackage{graphicx}
\usepackage{amsmath}
\usepackage{booktabs}
\usepackage{amssymb}
\usepackage{amsfonts}
\usepackage{multirow}

\title{\textsc{SymDiReC}: A Neuro-Symbolic Divide-Retrieve-Conquer Framework for Enhanced RTL Synthesis and Summarization}

\author{
 \textbf{Prashanth Vijayaraghavan},
 \textbf{Apoorva Nitsure},
 \textbf{Luyao Shi},
 \textbf{Charles Mackin},
\\
 \textbf{Ashutosh Jadhav},
 \textbf{David Beymer},
 \textbf{Ehsan Degan},
 \textbf{Vandana Mukherjee}
\\
\\
 IBM Research, San Jose, CA, USA
\\
\texttt{\{prashanthv,apoorva.nitsure,luyao.shi,charles.mackin\}@ibm.com}\\
\texttt{\{ashutosh,beymer,edehgha,vandana\}@us.ibm.com}
}

\begin{document}
\maketitle
\begin{abstract}
Register-Transfer Level (RTL) synthesis and summarization are central to hardware design automation but remain challenging for Large Language Models (LLMs) due to rigid HDL syntax, limited supervision, and weak alignment with natural language. Existing prompting and retrieval-augmented generation (RAG) methods have not incorporated symbolic planning, limiting their structural precision. We introduce \textbf{\textsc{SymDiReC}}\footnote{Short for Neuro-\textbf{Sym}bolic \textbf{Di}vide–\textbf{Re}trieve–\textbf{C}onquer Strategy}, a neuro-symbolic framework that decomposes RTL tasks into symbolic subgoals, retrieves relevant code via a fine-tuned retriever, and assembles verified outputs through LLM reasoning.  Supporting both Verilog and VHDL without LLM fine-tuning, \textsc{SymDiReC} achieves  $\sim$20\% higher Pass@1 rates for synthesis and 15–20\% \textsc{Rouge-L} improvements for summarization over prompting and RAG baselines, demonstrating the benefits of symbolic guidance in RTL tasks.
\end{abstract}

\section{Introduction}
\begin{figure}[t]
    \centering
    \includegraphics[width=0.475\textwidth]{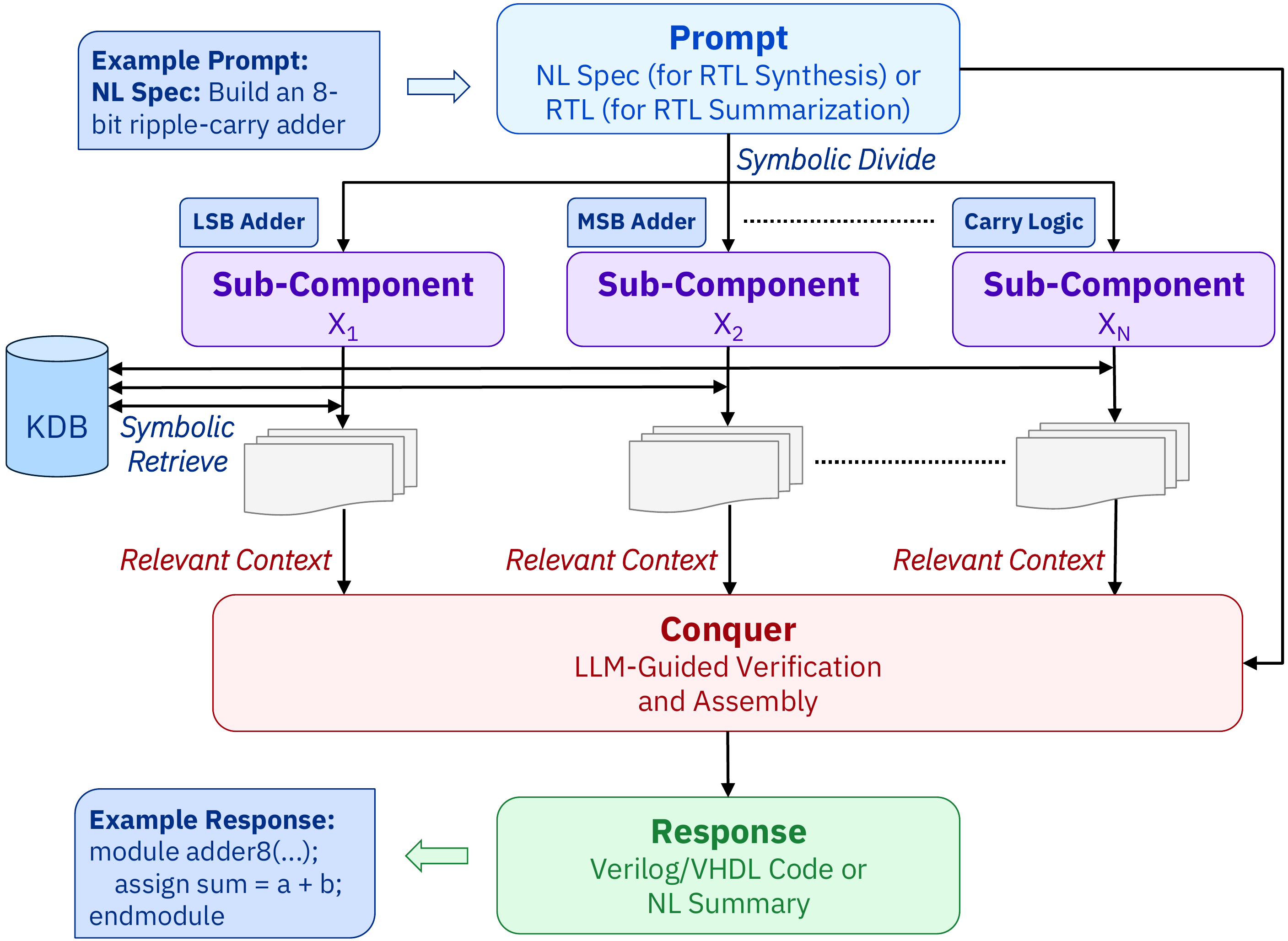}
\caption{Overview of our \textsc{SymDiReC} framework for RTL synthesis and summarization.}
    \label{fig:overview}
\end{figure}

Register-Transfer Level (RTL) synthesis and summarization are central tasks in Electronic Design Automation (EDA). RTL synthesis translates high-level natural language specifications into synthesizable hardware modules, while RTL summarization produces concise natural language explanations of existing hardware code. For example, given the natural language (NL) specification ``build an 8-bit ripple-carry adder'' a system must generate a correct Verilog/VHDL module that composes full-adders and propagates carries; conversely, given such a module, it should explain its functional structure (e.g., LSB/MSB adders and carry logic). While this example is relatively simple, real-world RTL designs often involve multi-stage pipelines, control logic, and hierarchical modules with complex timing and data dependencies. These characteristics make monolithic generation brittle and error-prone, motivating the need for modular decomposition, targeted retrieval of reusable RTL components, and explicit integration and verification. As illustrated in Figure~\ref{fig:overview}, both synthesis and summarization require preserving strict HDL syntax, modular structure, and precise functional semantics, distinguishing them from general-purpose code generation and summarization.


While Large Language Models (LLMs) have shown increasing promise in code generation, their performance on Hardware Description Languages (HDLs) like Verilog and VHDL remains limited due to rigid syntax, sparse annotated data, and semantic divergence from natural language \cite{vijayaraghavan2024vhdl, zhao2024codev}. Recent prompting strategies such as Chain-of-Thought (CoT) \cite{wei2022chain}, CoDes \cite{vijayaraghavan2024chain}, and ReAct \cite{yao2023react} enhance step-by-step reasoning yet struggle with RTL-specific tasks due to domain mismatch and absence of structural priors. Retrieval-Augmented Generation (RAG) methods \cite{lewis2020retrieval, petroni2020kilt, ho2024verilogcoder, ping2025hdlcore, yao2024hdldebugger} reduce hallucinations and improve factuality by grounding LLMs with external context. However, they often rely on expensive instruction-tuning, target only Verilog, and address either synthesis or summarization in isolation. Moreover, neither prompting nor RAG pipelines typically incorporate symbolic scaffolds, which can explicitly capture hardware intent.

Symbolic planning and neuro-symbolic approaches have demonstrated strong benefits in enhancing interpretability and structure-awareness in generation tasks \cite{zhou2022docprompting, pan2023logic}. Models like Logic-LM \cite{pan2023logic} and Code-as-Symbolic-Planner \cite{chen2025code} leverage symbolic scaffolding to guide generation, but these techniques have not been extended to RTL workflows. We introduce \textbf{\textsc{SymDiReC}}, a neuro-symbolic \textit{Divide–Retrieve–Conquer} framework tailored for RTL synthesis and summarization across both Verilog and VHDL. \textsc{SymDiReC} consists of three symbolic reasoning-driven stages: (a) \textbf{Divide} via symbolic decomposition, where an LLM breaks a high-level RTL task into modular sub-components with natural language and symbolic representations (e.g., Boolean or dataflow logic); (b) \textbf{Retrieve} using a domain-adapted symbolic retriever fine-tuned on the RTL-IR dataset, which incorporates both symbolic and textual cues to fetch semantically relevant RTL fragments; and (c) \textbf{Conquer} via LLM-guided verification and assembly, where retrieved candidates are aligned with symbolic intent and assembled into a final code block or summary. By integrating symbolic logic into every stage, \textsc{SymDiReC} improves retrieval precision, output consistency, and verification; all without requiring full LLM fine-tuning. Empirical results demonstrate that \textsc{SymDiReC} achieves roughly 20\% higher Pass@1 accuracy in synthesis and 15–20\% improvement in \textsc{Rouge-L} for summarization over strong RAG and prompting baselines. These results highlight the value of symbolic reasoning in bridging the gap between natural language, logic, and RTL semantics. Our key contributions are as follows:

\noindent \textbf{\textsc{SymDiReC} Framework:} We propose a novel neuro-symbolic Divide–Retrieve–Conquer pipeline for RTL tasks, integrating symbolic decomposition, retrieval, and verification in a unified architecture.

\noindent \textbf{Symbolic Guidance for Retrieval and Verification:} We show that symbolic logic enables more precise retrieval and structurally consistent outputs, outperforming natural language methods.

\noindent \textbf{Cross-Language RTL Evaluation:} We evaluate \textsc{SymDiReC} on both Verilog and VHDL benchmarks for synthesis and summarization, demonstrating generalizability across RTL domains.

\noindent \textbf{Lightweight and Modular Reasoning:} Our approach avoids full LLM fine-tuning by adapting only the retriever, maintaining competitive performance relative to several strong baselines.

\section{Related Work}

Transformer-based code models such as Megatron-LM \cite{shoeybi2019megatron}, StarCoder \cite{li2023starcoder}, CodeGen \cite{nijkamp2022codegen}, CodeLlama \cite{roziere2023code}, and Granite \cite{mishra2024granite} underpin much of the progress in multi-language generation and code reasoning. Prompting techniques, including Chain-of-Thought (CoT) \cite{wei2022chain}, CoDes \cite{vijayaraghavan2024chain}, and CoT with self-verification \cite{ping2025hdlcore}, deliver structured reasoning for hardware-related tasks. ReAct prompting \cite{yao2023react} adds iterative refine and act cycles to improve correctness further. Retrieval-Augmented Generation (RAG) grounds LLM outputs with external context \cite{lewis2020retrieval, petroni2020kilt}. Recent extensions such as self-reflective and corrective RAG \cite{asai2023self, yan2024corrective} and document-level prompting \cite{zhou2022docprompting} reduce hallucinations and streamline domain adaptation. Frameworks such as REDCODER \cite{parvez2021retrieval} have applied RAG for code summarization and generation in general-purpose languages, illustrating benefits of dual retrieval and generation.

In hardware design, modular RAG and reasoning strategies are emerging. VerilogCoder uses a task and circuit relation graph and AST-based debugging to exceed 90\% pass rates on Verilog benchmarks \cite{ho2024verilogcoder}. HDLCoRe and HDLdebugger add hardware-aware prompt decomposition plus evidence filtering \cite{ping2025hdlcore, yao2024hdldebugger}. ComplexVCoder implements a two-stage approach with intermediate representations and domain-specific RAG \cite{zuo2025complexvcoder}. Multi-level summarization models like CodeV \cite{zhao2024codev} improve Verilog generation via instruction tuning. All these methods rely on expensive model tuning or fine-tuning and typically focus only on Verilog or on one task, either synthesis or summarization.

Efforts specifically targeting VHDL are limited. The VHDL-Eval benchmark \cite{vijayaraghavan2024vhdl} and CoDes for VHDL \cite{vijayaraghavan2024chain} highlight consistent weaknesses of LLMs on VHDL, underscoring the need for methods that can handle both synthesis and summarization. Our approach, \textsc{SymDiReC}, uniquely addresses these gaps. It incorporates symbolic logic as a structured intermediate representation, enhancing both decomposition and retrieval. Unlike graph-only RAG frameworks, symbolic logic captures functional intent, allowing more precise retrieval and verification. \textsc{SymDiReC} is among the few systems evaluated on both Verilog and VHDL, and the first to jointly tackle synthesis and summarization across both languages. The neuro-symbolic combination composed of symbolic decomposition, retriever tuning for RTL semantics, and LLM-guided verification surpasses prior art in performance while maintaining interpretability and modular reasoning.

\begin{figure*}[t]
    \centering
    \includegraphics[width=0.95\linewidth]{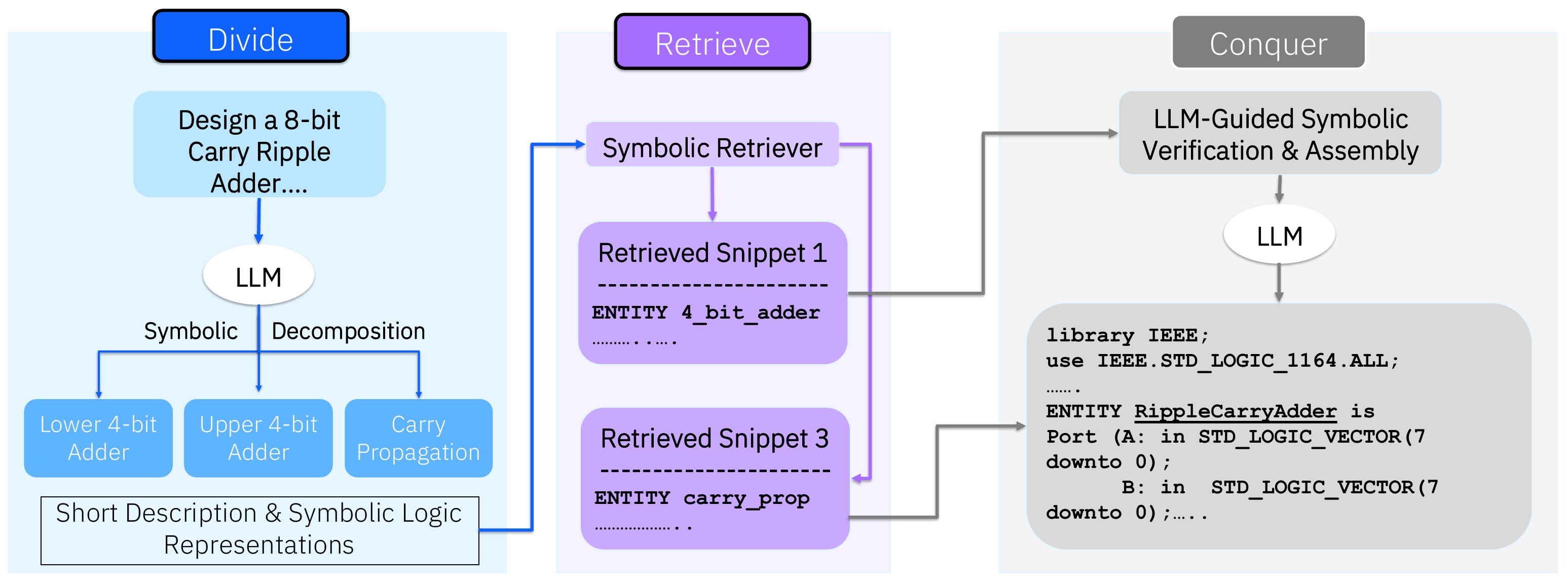}
\caption{Illustration of our \textsc{SymDiReC} framework for 8-bit carry ripple adder.}
    \label{fig:direc_overview}
\end{figure*}

\section{Neuro-Symbolic Divide-Retrieve- Conquer (\textsc{SymDiReC}) Framework}
\subsection{Overview}
We introduce \textbf{\textsc{SymDiReC}}, a unified neuro-symbolic framework designed for two complementary tasks in register-transfer level (RTL) design:  
(i) \textit{synthesis}, where natural language (NL) problem statements or specifications are translated into their corresponding RTL code; and  
(ii) \textit{summarization}, where RTL modules are converted into interpretable NL explanations augmented with symbolic logic.  By integrating symbolic reasoning into our pipeline, \textsc{SymDiReC} ensures semantically meaningful decomposition, retrieval, and verification. The symbolic representations act as an intermediate scaffold that enhances both retrieval precision and output correctness, especially in the context of more challenging RTL semantics. \textsc{SymDiReC} follows a three-stage architecture (refer Figure \ref{fig:direc_overview}) shared across both synthesis and summarization:
  \noindent \textbf{Divide (Symbolic Decomposition)}: The input, either a natural language specification or RTL code, is decomposed into smaller, semantically meaningful sub-components. Each sub-component is annotated with a brief textual description and a symbolic logic representation that captures its functional behavior.

  \noindent \textbf{Retrieve (Symbolic Querying)}: For each sub-component, the corresponding symbolic logic is used to retrieve relevant RTL snippets or symbolic summaries from a structured knowledge base. The retriever is trained to understand the alignment between symbolic logic and RTL structures.

  \noindent \textbf{Conquer (LLM Verification and Assembly)}: An LLM evaluates the retrieved candidates based on their alignment with the symbolic logic and description, selects the top candidate for each sub-component, and assembles them into a coherent RTL implementation or summary.

\subsection{Divide: Symbolic Decomposition}
\label{sec:decomp}

The \textbf{Divide} stage decomposes the input into a set of interpretable sub-components, each represented by a symbolic logic expression and a corresponding textual or structural unit. This decomposition step provides structure and granularity to the task, enabling downstream retrieval and verification at a finer granularity. Formally, for an input \(X\), the decomposition function is defined as:
\[
f_{\textsc{Div}}(X) = \{(x_1, \phi_1), \dots, (x_N, \phi_N)\}
\]
where \(x_i\) is the \(i^{\text{th}}\) sub-unit and \(\phi_i\) is its associated symbolic logic representation. The number of sub-units \(N\) is dynamic and may vary with the complexity of the input; it is governed by prompting strategies or syntax-driven partitioning mechanisms. For synthesis tasks, the input \(X\) is a natural language specification. We use a pretrained LLM to decompose this specification into a sequence of short NL descriptions \(\{x_i\}\), each denoting a distinct functional sub-task (e.g., counter, comparator, multiplexer). For each \(x_i\), we prompt the same LLM to produce a corresponding symbolic logic expression \(\phi_i\) that captures the intended behavior in a logic-based form (e.g., temporal, Boolean, or dataflow expressions). Optionally, we provide some in-context examples to encourage structured and consistent output formats grounded in symbolic hardware semantics. For summarization tasks, the input \(X\) is RTL code. We first parse the code into its abstract syntax tree (AST) and segment it into functional blocks \(\{x_i\}\), such as `always` blocks, modules, or combinational logic segments. For each block \(x_i\), we prompt an LLM to abstract its functional intent into a concise symbolic representation \(\phi_i\), typically capturing control logic, state transitions, or data transformations. We encourage structural consistency in these symbolic forms using prompt-based templates grounded in RTL semantics. 

\textbf{LLM Guidance and Symbolic Abstraction:} In both directions, symbolic abstraction relies on prompting a pretrained LLM to reason about hardware functionality and generate interpretable symbolic expressions. While LLMs may not always produce complete or formal logic, their output often provides high-quality approximations that capture key semantic elements of the underlying sub-component. These symbolic sketches serve as anchors for downstream retrieval and alignment. This decomposition process ensures that each sub-unit is semantically meaningful and structurally aligned with RTL design principles, enabling targeted retrieval and robust composition in later stages.

\subsection{Retrieve: Symbolic Retriever}
\label{sec:retrieval}

The \textbf{Retrieve} stage enriches each sub-component \((x_i,\phi_i)\) with relevant RTL code snippets or symbolic/NL summaries. To leverage both the textual description \(x_i\) and its formal symbolic logic \(\phi_i\), we design a joint embedding retriever that matches paired queries \((x_i,\phi_i)\) against a structured knowledge base \(\mathcal{K}\). We denote them as:
\[
  f_{\textsc{Ret}}\bigl(x_i,\phi_i\bigr) = R_i
  \;=\;\mathrm{TopK}_{y\in\mathcal{K}}\;\mathrm{score}(x_i,\phi_i;y).
\]

\subsubsection{Knowledge Base}
We construct a repository of \(S\) indexed entries:
$
\mathcal{K}=\bigl\{(y_j,d_j,\phi_j)\bigr\}_{j=1}^S,
$
where each entry contains: (a) \(y_j\): an RTL code snippet (VHDL/Verilog) or NL summary, (b) \(d_j\): a short NL explanation of \(y_j\), and (c) \(\phi_j\): symbolic logic representation.

\subsubsection{Joint Retriever Architecture}
We implement a dual‐encoder with three transformer encoders:
$
e_x: \mathcal{X}\to\mathbb{R}^D,\quad
e_\phi: \Phi\to\mathbb{R}^D,\quad
e_y: \mathcal{Y}\to\mathbb{R}^D,
$
where \(e_x\) embeds NL fragments \(x\), \(e_\phi\) embeds symbolic logic \(\phi\), and \(e_y\) embeds candidate entries \(y\).  To form a joint query representation, we concatenate and project:
\[
q_i = W_q\bigl[e_x(x_i)\,\Vert\,e_\phi(\phi_i)\bigr]\;\in\;\mathbb{R}^D,
\]
with learned projection matrix \(W_q\in\mathbb{R}^{D\times2D}\).  Retrieval then ranks each entry \(y_j\) by cosine similarity:
\[
\mathrm{score}(x_i,\phi_i;\,y_j) = \cos\bigl(q_i,\;e_y(y_j)\bigr).
\]

\subsubsection{Training Objective}
We fine‐tune all three encoders on our \textsc{Rtl-Ir} dataset of aligned triplets \(\{(x_p,\phi_p,y_p)\}_{p=1}^S\).  In each batch of size \(B\), the positive example \((x_p,\phi_p,y_p)\) is contrasted against in‐batch negatives \(\{y_q\}_{q\neq p}\).  We minimize the multiple‐negatives ranking loss and explore both dense (continuous embeddings) and sparse (term‐weighted) variants for \(e_x\) and \(e_\phi\); implementation and hyper‐parameter details appear in the Appendix \ref{sec:retriever}, along with dataset statistics and ablations.

\begin{table*}[t]
\centering
\small

\begin{tabular}{@{}llcccc@{}}
\toprule
\textbf{Method} & \textbf{LLM} & \multicolumn{2}{c}{\textbf{Pass@1}} & \multicolumn{2}{c}{\textbf{\textsc{Rouge‑L}}} \\
 &  & \textbf{Verilog} & \textbf{VHDL} & \textbf{Verilog} & \textbf{VHDL} \\
\midrule

\multirow{2}{*}{Vanilla Prompting}    & GPT‑4o  & $0.543$ & $0.285$ & $43.1$ & $39.3$ \\
                                       & Llama‑3 & $0.385$ & $0.226$ & $40.2$ & $34.1$ \\
\midrule

\multirow{2}{*}{CoDes} & GPT‑4o  & $0.602$ & $0.348$ & $46.9$ & $43.2$ \\
                                       & Llama‑3 & $0.435$ & $0.274$ & $43.5$ & $39.5$ \\
\midrule

\multirow{2}{*}{ReAct Prompting}       & GPT‑4o  & $0.616$ & $0.353$ & $46.1$ & $43.0$ \\
                                       & Llama‑3 & $0.437$ & $0.291$ & $42.9$ & $38.8$ \\
\midrule

\multirow{2}{*}{VRAG‑CodeBERT}         & GPT‑4o  & $0.688$ & $0.487$ & $53.2$ & $50.3$ \\
                                       & Llama‑3 & $0.527$ & $0.396$ & $47.4$ & $44.5$ \\
\midrule

\multirow{2}{*}{VRAG‑FT}               & GPT‑4o  & $0.719$ & $0.531$ & $57.0$ & $52.8$ \\
                                       & Llama‑3 & $0.569$ & $0.439$ & $50.5$ & $48.1$ \\
                                       
\midrule

\multirow{2}{*}{RTLCoder (open‑source)} & Mistral & \(0.625^{\ast}\) & - & - & - \\
 & GPT-4o/Llama‑3 & - & - & - & - \\
\midrule

\multirow{2}{*}{CodeV (instruction‑tuned)} & CodeQwen & \(0.532^{\ast}\) & - & - & - \\
 & GPT-4o/Llama‑3 & - & - & - & - \\ \midrule

\multirow{2}{*}{\textsc{SymDiReC} (ours)} 
& GPT‑4o  & $\mathbf{0.805}_{\pm 0.020}$ & $\mathbf{0.634}_{\pm 0.022}$ & $\mathbf{62.5}_{\pm 0.015}$ & $\mathbf{56.6}_{\pm 0.018}$ \\
& Llama‑3 & $\mathbf{0.652}_{\pm 0.022}$ & $\mathbf{0.545}_{\pm 0.020}$ & $\mathbf{56.1}_{\pm 0.018}$ & $\mathbf{50.8}_{\pm 0.015}$ \\
\midrule
\midrule

\multirow{2}{*}{\textsc{SymDiReC}‑GT (oracle)}   & GPT‑4o  & $\textbf{0.902}$ & $\textbf{0.842}$ & $\textbf{70.2}$ & $\textbf{64.7}$ \\
                                       & Llama‑3 & $\textbf{0.807}$ & $\textbf{0.721}$ & $\textbf{63.3}$ & $\textbf{57.9}$ \\

\bottomrule
\end{tabular}
\caption{RTL synthesis (Pass@1) and summarization (\textsc{Rouge-L}) performance across methods and LLMs. For our \textsc{SymDiReC}, mean $\pm$ standard deviation over five independent runs is reported. Results are statistically significant vs. the strongest baseline (paired t‑test, \(p < 0.01\)). $^*$ indicates results reported in the original papers.}
\label{tab:rtl-results}
\end{table*}

\subsubsection{Inference: Retrieving Sub-components}
\label{sec:retrieve-inference}

At inference time, each decomposed query \((x_i,\phi_i)\) is encoded as \( q_i \) and we compute the cosine similarity score between the computed query and candidate \(y_j\in\mathcal{K}\). The retriever returns the top‑\(k\) candidates as:
\[
R_i = f_{\textsc{Ret}}(x_i,\phi_i)
= \mathrm{TopK}_{j\in[1,S]}\,\mathrm{score}(x_i,\phi_i;y_j),
\]
yielding \(R_i=\{r_{i,m}\}_{m=1}^k\). Each \(r_{i,j}\) is either an RTL snippet (for synthesis) or a NL summary (for summarization). Table~\ref{tab:qualitative-rippleadder-symbolic} presents a qualitative example of an 8-bit ripple-carry adder, including symbolic decompositions into Boolean/logical expressions for each submodule and the retrieved Verilog and VHDL code snippets tied to these submodules.

\subsection{Conquer with LLM-Guided Verification and Assembly}
\label{sec:conquer}

The \textbf{Conquer} stage integrates retrieved candidates into a finalized output. Given the original input \(X\) and the retrieved sets \(\{R_i\}_{i=1}^N\), the final artifact \(\hat Y\) is produced by:
$
\hat{Y} \;=\; f_{\textsc{Conq}}\bigl(X,\{\!R_i\!\}_{i=1}^N\bigr),
$
where \(\hat Y\) is either the synthesized RTL module or the summarized NL description. Given \(R_i=\{r_{i,m}\}_{m=1}^k\) are the top‑\(k\) candidates for each sub-component \(i\),
 from Section~\ref{sec:retrieval}, we prompt the generator LLM to assign an alignment score by conditioning on the sub-component $x_i$ and its associated symbolic logic representation $\phi_i$ \; $\forall m\in\{1,k\}$ as:
$
\hat{\alpha}_{i,m}
= \mathrm{verify\_score}\bigl(r_{i,m},\,x_i,\,\phi_i\bigr)
\;\in[0,1]; 
$
These scores reflect both functional correctness and fidelity of retriever results to the symbolic specification. We select the highest‑scoring candidate for each sub-component. This yields a verified set of sub-modules or summaries \(\{\hat{r}_i\}_{i=1}^N\).
The final assembly invokes the LLM conditioned on \(X\) and the verified candidates. This step produces a coherent RTL design or comprehensive NL summary, ensuring consistent naming, module connectivity, and logical integrity. 



\medskip
\noindent

\section{Experiments}
This section outlines our experimental setup, including RTL benchmarks (in both VHDL and Verilog), a diverse suite of baseline models, and evaluation metrics designed to assess the effectiveness, generalization, and efficiency of the proposed \textsc{SymDiReC} framework. Refer Appendix~\ref{app:datasets} for full implementation and dataset details. Our evaluation is structured around the following research questions: \textbf{(RQ1) Effectiveness of \textsc{SymDiReC}:} How effective is the \textsc{SymDiReC} framework for RTL synthesis and summarization, compared to existing baseline approaches?  \textbf{(RQ2) Impact of Symbolic Logic:} To what extent do symbolic logic representations improve retrieval quality and downstream RTL generation or summarization? \textbf{(RQ3) Hyperparameter Sensitivity:} How do key hyperparameters affect the performance of \textsc{SymDiReC}?

\begin{figure*}[!ht]
    \centering
    \includegraphics[width=0.98\textwidth]{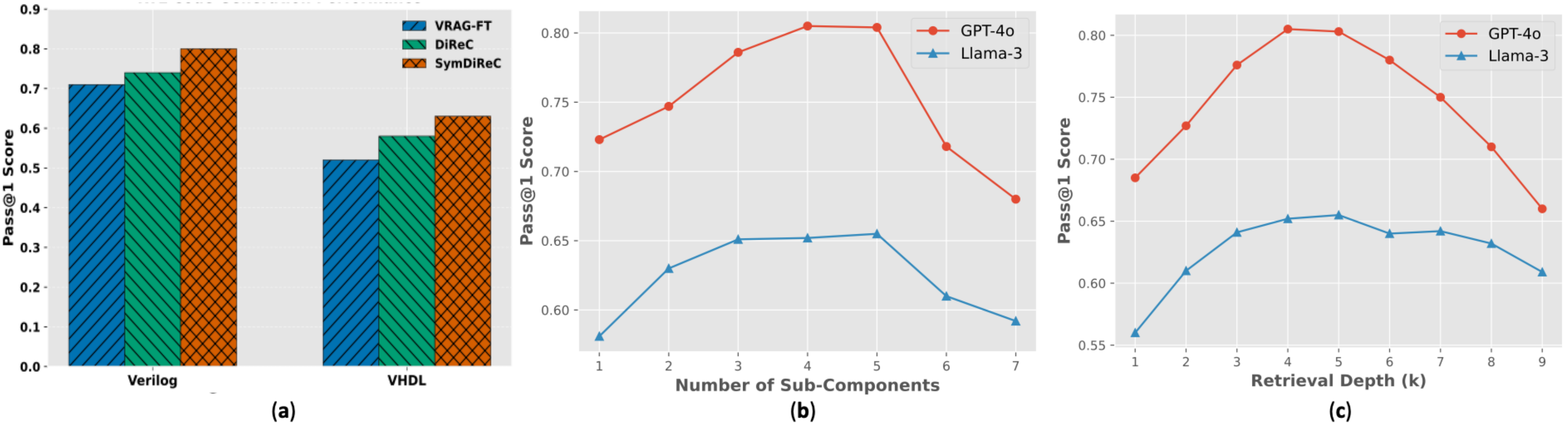}
\caption{Ablation results: Performance with varying (a) number of sub-components and (b) chunking strategy.}
\label{fig:ablations}
\end{figure*}

\subsection{Benchmarks and Baselines}

We evaluate \textsc{SymDiReC} on two standard RTL benchmarks: \textbf{Verilog-Eval}, comprising 156 functional Verilog tasks from HDLBits~\cite{liu2023verilogeval} with testbenches, and \textbf{VHDL-Eval}, containing 202 VHDL tasks translated from Verilog-Eval or public tutorials~\cite{vijayaraghavan2024vhdl}, with analogous verification procedures. Our comparisons include zero-shot prompting, intermediate plan-based CoDes~\cite{vijayaraghavan2024chain}, iterative ReAct~\cite{yao2023react}, Vanilla RAG (VRAG-CodeBERT), and RAG fine-tuned on the \textbf{RTL-IR} dataset (VRAG-FT). We also include recent domain-specialized models RTLCoder~\cite{liu2024rtlcoder} and CodeV~\cite{zhao2024codev}. Finally, we evaluate \textsc{SymDiReC} and its oracle variant using ground-truth snippets (\textsc{SymDiReC}-GT). The RTL-IR dataset is a curated collection of RTL code and annotations used to finetune RAG models, comprising text-to-code, functionally equivalent code (FEC), code-to-summary, and partial-to-complete code pairs. Table~\ref{tab:stats} summarizes the dataset statistics. Detailed dataset descriptions, benchmarks, and additional examples are provided in Appendix~\ref{app:datasets}, \ref{app:benchmarks}. Baselines use GPT-4o and Llama-3 (70B), evaluated with Pass@1 for synthesis and \textsc{Rouge-L} for summarization. Results report mean $\pm$ std over 5 runs with paired t-tests.


\begin{table}[]
\small
\centering
\begin{tabular}{@{}lr@{}}
\toprule
\multicolumn{2}{c}{RTL-IR Dataset Statistics} \\ \midrule
\# Total Size                     & $\sim 50.5$k  \\
\# Text-to-Code Pairs             & $\sim 8$k     \\
\# FEC Pairs                      & $\sim 13.5$k  \\
\# Code-to-Summary Pairs          & $\sim 6.5$k   \\
\# Partial-to-Complete Code Pairs & $\sim 22.5$k  \\ \bottomrule
\end{tabular}
\caption{Dataset statistics of RTL-IR used for model finetuning and retrieval enhancement.}
\label{tab:stats}
\end{table}

\subsection{Evaluation Metrics}
\label{sec:metrics}
For RTL synthesis, we use the metric {Pass@1}, which denotes the proportion of first-attempt RTL designs that successfully pass the self-checking testbenches provided in the Verilog‑Eval and VHDL‑Eval benchmarks. For code summarization, we employ \textsc{Rouge‑L}, measuring the longest common subsequence (LCS) between generated and reference summaries and computing an F-measure to assess fluency and coherence \cite{Lin:2004:ROUGE}.

\section{Results \& Discussion}
\subsection{Overall Performance of \textsc{SymDiReC}}
Table~\ref{tab:rtl-results} compares \textsc{SymDiReC} against strong baselines, including prompting-only methods and retrieval-augmented generation (RAG) strategies. Our neuro-symbolic pipeline consistently outperforms all alternatives, demonstrating the effectiveness of symbolic decomposition, a domain-adapted retriever, and LLM-guided verification.

\subsection{Effectiveness of \textsc{SymDiReC} (RQ1)}
Vanilla Prompting serves as the zero-shot lower bound for each model, showing the weakest performance across both synthesis and summarization tasks. In contrast, \textsc{SymDiReC}-GT represents an approximate upper bound for RAG-based methods, as it always includes the ground-truth among the top-$k$ retrieved candidates. Despite having the correct solution in context, the failure of \textsc{SymDiReC}-GT in specific experimental cases indicates that the LLM can still struggle to filter out distractors within the retrieved candidates or may lack sufficient RTL-specific reasoning capabilities. Chain-of-Descriptions (CoDes) and ReAct prompting yield comparable performance. ReAct demonstrates a modest edge ($\sim${5-8}\%) relative improvement in Pass@1, likely because its iterative reasoning/action loop allows correction pathways that simplistic descriptive chains lack. Among RAG strategies, VRAG-FT with a fine-tuned retriever consistently outperforms VRAG-CodeBERT, achieving $\sim$10\%-15\% relative improvement in Pass@1 and $\sim$5\%-8\% in \textsc{Rouge‑L}. This suggests that while generic code retrievers benefit RTL tasks, task-specific tuning further enhances retrieval quality by aligning natural language queries more effectively with RTL semantics. \textsc{SymDiReC} outperforms all baselines by significant margins: up to $\sim$80\% relative improvement over the best prompt-based method and up to $\sim$20\% over RAG-only baselines in Pass@1. For summarization, \textsc{SymDiReC} yields up to $\sim$35\% gains over prompting and $\sim$10\% over RAG in \textsc{Rouge-L}, highlighting the benefits of symbolic planning and RTL-aware retrieval; the remaining $\sim$10--15\% gap to \textsc{SymDiReC}-GT indicates room to improve retriever precision and LLM alignment.


\subsection{Impact of Symbolic Logic (RQ2)}

To isolate the benefit of symbolic logic, we compare three pipeline variants: VRAG‑FT, which applies a fine-tuned retriever on NL queries; DiReC, which uses the same Divide–Retrieve–Conquer structure but uses NL-only queries for retrieval; and our full \textsc{SymDiReC}, which incorporates symbolic decomposition with modular retrieval and LLM-guided verification. We find that \textsc{SymDiReC} achieves $\sim${10-15\%} relative improvements in Pass@1 (refer Figure~\ref{fig:ablations}(a)) and \textsc{Rouge-L} versus both VRAG‑FT and DiReC. These findings demonstrate that symbolic representations serve as a strong scaffolding mechanism, enabling more precise retrieval, reducing noise, and thereby enhancing both synthesis and summarization outcomes.


\subsection{Hyperparameter Sensitivity (RQ3)}
We perform ablations to explore how key hyperparameters, namely the number of sub-components ($N$) and retrieval depth ($k$), affect \textsc{SymDiReC}'s performance. Figures~\ref{fig:ablations}(b) and (c) summarize the results. Increasing $N$ from 2 to 6 leads to consistent improvements in Pass@1; however, performance starts to decline when $N$ exceeds 6, likely due to excessive fragmentation that results in semantically weaker sub-units. Importantly, the optimal value of $N$ is \emph{task-dependent}. Simpler combinational tasks (e.g., adders, multiplexers) benefit from smaller decompositions ($N\approx3$--$4$), while multi-stage sequential designs (e.g., counters, FSMs) achieve better performance with slightly larger $N$ ($N\approx4$--$5$), reflecting their increased structural complexity. For retrieval depth, performance improves as $k$ increases up to 5, but plateaus and eventually drops when $k$ becomes too large. This decline is likely caused by additional noise introduced by less relevant retrievals. Across benchmarks, a default configuration of $N=4$ and $k=5$ provides a strong balance between synthesis accuracy, summarization quality, and computational efficiency, while allowing task-specific tuning when appropriate.

\section{Error Analysis}
\label{app:error_analysis}

We perform a detailed error analysis to understand the limitations of \textsc{SymDiReC} in RTL synthesis and summarization. Our investigation focuses on three major sources of errors: (a) \textbf{Symbolic Decomposition Errors:} Approximately 8-10\% of sub-components have incomplete or inconsistent symbolic expressions, particularly for multi-bit comparators or sequential elements. These errors correlate with lower retrieval precision, reducing Pass@1 performance by up to 5-7\% for affected designs; (b) \textbf{Retrieval Mismatches:} Even with symbolic scaffolds, around 12-15\% of retrieved candidates only partially match the intended behavior or contain distractors. This results in a 3-6\% drop in Pass@1 accuracy and 2-4 \textsc{Rouge‑L} points in summarization; and (c) \textbf{LLM Assembly \& Verification Failures:} When retrieved candidates are correct, the LLM occasionally fails to integrate them properly (signal misalignment, missing connections, or carry propagation issues), observed in roughly 6-8\% of sub-components. These failures contribute to a remaining gap of $\sim$10-15\% between \textsc{SymDiReC} and the oracle \textsc{SymDiReC}-GT in synthesis and 6–8 \textsc{Rouge‑L} points in summarization. Qualitative inspection shows that most failures occur in hierarchical designs or uncommon module patterns. The \textsc{SymDiReC}-GT results suggest that improved retrieval precision and symbolic reasoning could close a significant portion of the performance gap.

\section{Conclusion}

We presented \textsc{SymDiReC}, a neuro-symbolic Divide–Retrieve–Conquer framework for RTL synthesis and summarization across Verilog and VHDL. By integrating symbolic decomposition, domain-adapted retrieval, and LLM-guided verification, \textsc{SymDiReC} effectively bridges the gap between formal hardware semantics and large language model generation. Unlike prior approaches that rely heavily on instruction tuning or overlook symbolic intent, our method introduces structured intermediate reasoning to improve both retrieval relevance and generation correctness. Empirical results demonstrate consistent improvements over prompting- and RAG-based baselines, with gains in synthesis accuracy and summarization quality. This work underscores the utility of symbolic planning in program synthesis and opens new directions for interpretable and modular neuro-symbolic systems in hardware design automation and beyond.

\section*{Limitations}
While \textsc{SymDiReC} demonstrates strong performance in RTL synthesis and summarization, several limitations remain. Symbolic decomposition depends on the LLM's ability to generate well-formed symbolic expressions. Smaller or less capable models may produce incomplete or inconsistent decompositions, diminishing the symbolic scaffolding benefits and yielding performance similar to natural language-only queries. The LLM-guided verification in the \textsc{Conquer} stage can also fail to align retrieved candidates with the intended logic, particularly when top-k retrievals include distractors or partially matching snippets. Decomposition granularity is sensitive: overly fine segmentation fragments the input, producing weak sub-units, while overly coarse segmentation may reduce retrieval precision. The current pipeline is restricted to single-file RTL designs and does not support hierarchical or multi-file projects, where cross-module dependencies are common. Scaling to such designs may require advanced AST processing, block- and function-level chunking, and multi-level summarization strategies. Even with the correct solution retrieved, the LLM may fail to select it, highlighting challenges in noise filtering, candidate ranking, and reasoning under imperfect retrieval. Addressing these issues: improving symbolic reasoning, retrieval alignment, and hierarchical abstraction, will be essential to extending \textsc{SymDiReC} to complex real-world hardware design scenarios.

\section*{Ethics Statement}

We use publicly available datasets (e.g., Verilog-Eval) and our curated RTL-IR dataset, which is sourced from permissively licensed GitHub repositories (MIT, BSD, Apache-2.0); license metadata is provided in the supplementary material. No private or sensitive data was used; outputs are intended for research and developer-assist purposes only. Potential risks include generating hardware designs that may be incorrect or unsafe if deployed without verification. Our system is intended as a developer-assist tool, and all outputs should be validated using standard testbenches and human review before real-world use.

\bibliography{main}

\newpage
\appendix

\section{Datasets}
\label{app:datasets}
\subsection{\textsc{RTL-IR}}

\subsubsection{Data Collection and Preprocessing}
The dataset was curated from publicly available VHDL/Verilog repositories on GitHub. We filtered repositories based on permissive licensing and selected VHDL/Verilog projects with meaningful comments and README descriptions. The preprocessing pipeline involved:
\begin{itemize}
    \item Extracting comments and README documentation.
    \item Using in-context learning (ICL) with Granite-13b-Instruct to refine problem statements and code summaries.
    \item Applying various transformations to generate functionally equivalent code pairs.
\end{itemize}

\paragraph{Text-to-Code (TC) Pairs} 
To construct TC pairs, we extracted natural language descriptions from comments in VHDL/Verilog files and relevant portions of README documentation. Since raw comments may be unstructured, ICL with Granite-13b-Instruct was used to generate structured problem statements. These statements were validated to ensure clarity and relevance to the corresponding VHDL/Verilog code. 

\paragraph{Code-to-Summary (CS) Pairs}
CS pairs were created by mapping VHDL/Verilog code to textual summaries. Code files with well-commented structures were prioritized, and ICL was employed to convert detailed comments into concise summaries. To assess summary quality, we manually annotated 100 examples, classifying them into:
\begin{itemize}
    \item \textit{Good} (clear, precise, and informative).
    \item \textit{Acceptable} (partially informative but useful).
    \item \textit{Bad} (incomplete or misleading).
\end{itemize}
Overall, 84\% of summaries were classified as \textit{good} or \textit{acceptable}, while 16\% were \textit{bad}. The latter were treated as ``hard negatives.''

\paragraph{Functionally Equivalent Code (FEC) Pairs} 
FEC pairs were generated by applying different transformation strategies to create variations of functionally identical VHDL/Verilog code. The transformations include:

\begin{itemize}
    \item \textbf{Type-2:} Renaming identifiers while maintaining functional equivalence. We extracted and renamed entity, architecture, process, and port names using an LLM-based renaming strategy. Single-character identifiers were replaced with LLM-suggested alternatives, while complex identifiers underwent abbreviation, permutation, or transformation to maintain readability.
    \item \textbf{Type-3:} Modifying statement order and introducing functionally inert code, ensuring variation while preserving functionality. Reordering declarations and restructuring conditional logic introduced additional diversity.
    \item \textbf{Type-4:} Back-translation between VHDL and Verilog using GHDL and ICARUS Iverilog. This process altered variable names and introduced intermediate signals, capturing functionally equivalent structures while minimizing lexical similarities.
\end{itemize}

\begin{figure*}[!ht]
    \centering
    \includegraphics[width=0.95\textwidth]{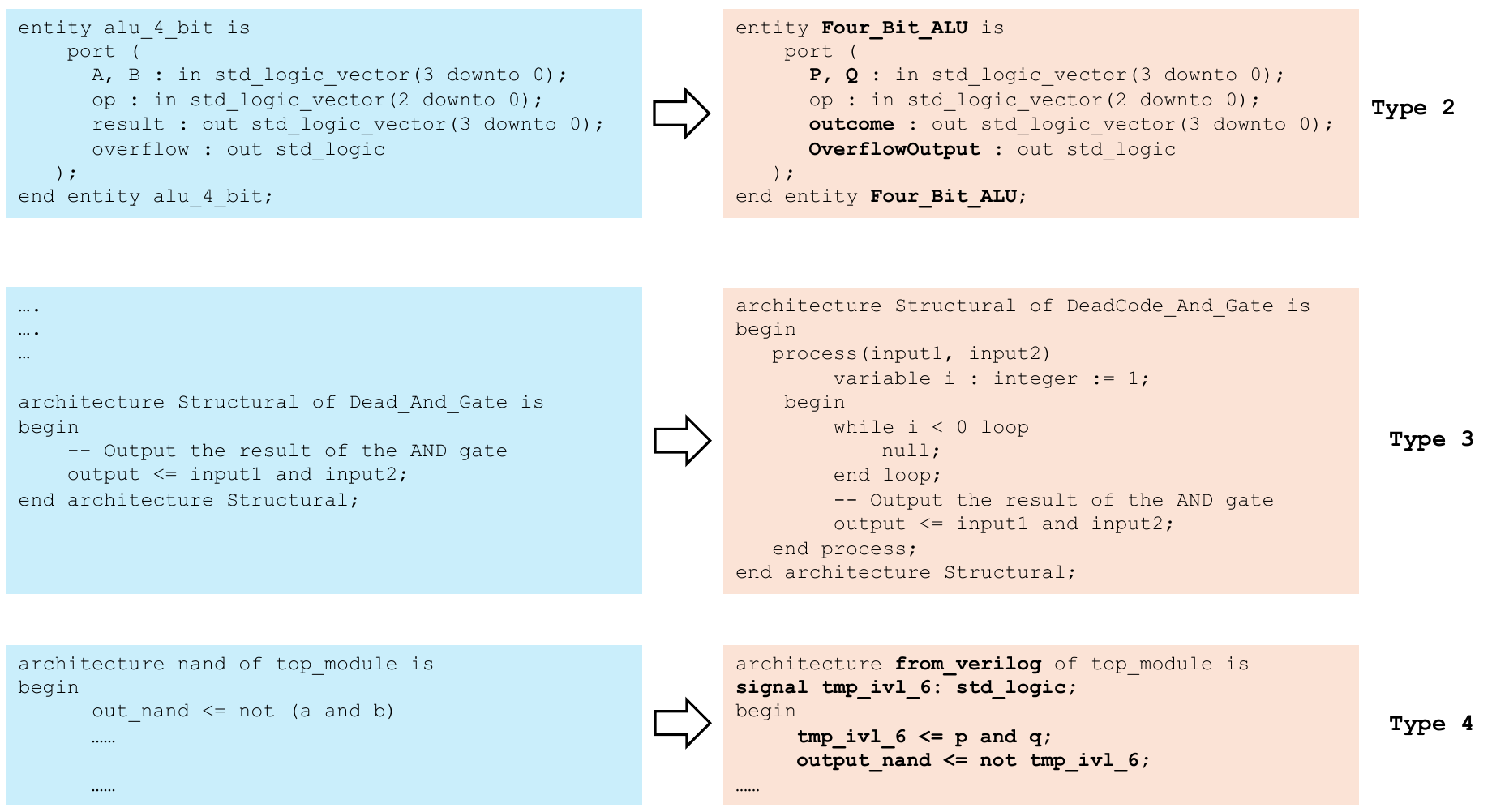}
\caption{VHDL Samples of different transformation strategies applied using the three categories of code clones -- Type 2, Type 3 and Type 4.}
    \label{fig:clone_types}
\end{figure*}

Figure~\ref{fig:clone_types} illustrates these transformation types with examples.


\paragraph{Partial-to-Complete Code (PC) Pairs} 
PC pairs were created by extracting partial VHDL/Verilog snippets from larger codebases and pairing them with their complete versions. To ensure lexically diverse representations, Type-2 transformations were applied to a subset of the complete versions. Snippet extraction was limited to code sections containing fewer than 1024 tokens, capturing function declarations and entity definitions along with relevant contextual comments.

\subsubsection{Quality Control Measures}
To ensure dataset integrity and usefulness, the following quality control measures were applied:
\begin{itemize}
    \item \textbf{Compilation Validation:} All functionally transformed code underwent compilation tests to ensure correctness.
    \item \textbf{Testbench Execution:} Available testbenches from GitHub were executed to verify functional equivalence.
    \item \textbf{Manual Review:} Code summaries were manually reviewed, with low-quality summaries marked as ``hard negatives.''
\end{itemize}

These measures enhance dataset reliability, ensuring it serves as a strong benchmark for VHDL code generation and summarization tasks.

\section{Training \& Evaluation of Retriever}
\label{sec:retriever}
\noindent\textbf{Retrieval} 
We fine-tune models from three categories: sparse, dense, and hybrid retrievers, using the RTL-IR dataset. The sparse retriever is BM25 \cite{robertson2009probabilistic}. For dense models, we fine-tune top performers from the MTEB Leaderboard \cite{muennighoff2022mteb} (GTE-Qwen-1.5b \cite{li2023towards}, Stella-400m\cite{zhang2023stella}, GIST-Large \cite{solatorio2024gistembed}), as well as CodeT5+ \cite{wang2021codet5} and Sentence Transformer (ST) \cite{reimers2019sentence} for code embeddings. Hybrid methods include SPLADE \cite{formal2021splade} and ELSER\cite{elasticML2024}. All, except ELSER, are fine-tuned on the RTL-IR training set. We evaluate on the held-out test set.

\subsection{Evaluation Metric}
\label{sec:ndcg_metrics}

\noindent\textbf{Retrieval:} We employ Normalized Discounted Cumulative Gain (NDCG) to assess ranking quality, rewarding highly relevant results appearing earlier in the list. NDCG@1 measures the relevance of the top-ranked result, while NDCG@10 evaluates ranking effectiveness across the top 10 positions, assigning higher weights to top-ranked items.

\subsection{Performance of Retrievers}
\label{sec:rq1}
Table \ref{tab:vhdl_irr} presents the performance of various retrieval methods on the \textsc{RTL-IR} test set using NDCG@1 and NDCG@10 metrics. The results indicate that dense retrieval methods consistently outperform hybrid and sparse approaches, as \textsc{RTL-IR} requires identifying semantically relevant matches beyond surface-level lexical overlaps. Among the evaluated models, CodeT5+ achieves the highest performance, with an NDCG@1 of 0.657 and an NDCG@10 of 0.872, demonstrating its strong ability to retrieve relevant VHDL/Verilog code snippets. This performance advantage can be attributed to CodeT5+'s pre-training on VHDL/Verilog code before fine-tuning on \textsc{RTL-IR}. 

Text embedding models such as Stella-400m (NDCG@1 = 0.656) and GTE-Qwen-1.5b (NDCG@1 = 0.644) follow closely, despite being fine-tuned solely on the \textsc{RTL-IR} dataset. Their effectiveness is linked to their large parameter and embedding sizes (e.g., Stella-400m with an embedding size of 8192), enabling better generalization. However, CodeT5+'s code-specific training appears to compensate for its smaller embedding size, leading to superior retrieval performance. 

\begin{table}[]
\centering
\small

\begin{tabular}{@{}lcc@{}}

\toprule
\textbf{Methods} & \textbf{NDCG@1} & \textbf{NDCG@10} \\ \midrule
BM25             & 0.434           & 0.570            \\
ELSER            & 0.485           & 0.664            \\
SPLADE           & 0.577           & 0.688            \\
GTE-Qwen-1.5b    & 0.644           & 0.864            \\
Stella-400m      & 0.656           & 0.866            \\
GIST-Large       & 0.616           & 0.802            \\
CodeT5+          & \textbf{0.657}  & \textbf{0.872}   \\
ST               & 0.556           & 0.665            \\ \bottomrule
\end{tabular}
\caption{Evaluation of sparse, hybrid, and dense retrievers on \textsc{RTL-IR} test set.}

\label{tab:vhdl_irr}
\end{table}

 \section{Benchmarks and Baselines}
\label{app:benchmarks}

\subsection{Benchmarks}

\textbf{Verilog-Eval:} 156 functional Verilog tasks sourced from HDLBits~\cite{liu2023verilogeval}, each with a self-verifying testbench. The tasks span a variety of RTL constructs, including combinational logic, sequential modules, counters, comparators, and simple finite-state machines.

\textbf{VHDL-Eval:} 202 VHDL tasks translated from Verilog-Eval problems or drawn from public VHDL tutorials~\cite{vijayaraghavan2024vhdl}, also with testbenches for functional verification. The suite maintains a similar functional diversity to Verilog-Eval.

\subsection{Baselines}

We compare \textsc{SymDiReC} against several baselines, including recent domain-specialized models:

\noindent \textbf{Vanilla Prompting (ZS):} Zero-shot, natural language to RTL code / summary generation.

\noindent \textbf{Chain-of-Descriptions (CoDes)}~\cite{vijayaraghavan2024chain}: Uses intermediate textual plans (descriptions) to guide LLM synthesis.

\noindent\textbf{ReAct Prompting}~\cite{yao2023react}:Iterative reasoning-and-action loops for stepwise generation and refinement.

\noindent \textbf{Vanilla RAG (VRAG-CodeBERT):} Uses a generic CodeBERT retriever without RTL-specific fine‑tuning.

\noindent \textbf{VRAG-FT:} RAG with a retriever fine-tuned on our RTL-IR dataset of aligned (NL, symbolic logic, RTL) triplets.

\noindent \textbf{RTLCoder}~\cite{liu2024rtlcoder}: An open-source LLM trained specifically for RTL generation, designed to be efficient and locally deployable.  

\noindent \textbf{CodeV}~\cite{zhao2024codev}: Instruction-tuned Verilog generation LLMs using a multi-level summarization strategy, shown to improve code generation by first summarizing then generating.  

\noindent \textbf{\textsc{SymDiReC}:} Our full neuro-symbolic pipeline, combining symbolic decomposition, retrieval, and LLM-guided verification.  

\noindent \textbf{\textsc{SymDiReC}-GT (Oracle):} An upper‑bound variant that retrieves using ground-truth symbolic snippets, to assess ideal retrieval conditions.

\subsection{LLM Settings and Evaluation}

We run all methods using two LLMs: a proprietary model (GPT‑4o) and an open-source model (Llama‑3, 70B). Synthesis correctness is measured via self-verifying testbenches using {Pass@1}, and summarization quality is scored using \textsc{Rouge-L} against reference summaries. For statistical robustness, we perform five independent runs per setting and report mean and standard deviation; paired t‑tests (e.g., comparing \textsc{SymDiReC} vs VRAG‑FT or vs other baselines) are computed and reported in the main results table.

\begin{table*}[ht]
\centering
\small
\begin{tabular}{@{}lp{5cm}p{7cm}@{}}
\toprule
\textbf{Task} & \textbf{Symbolic Decomposition} & \textbf{Retrieved Context} \\
\midrule
8-bit Ripple Carry Adder & 
\textbf{LSB Half-Adder:} $S_0 = A_0 \oplus B_0$, $C_1 = A_0 \land B_0$ \newline
\textbf{Bits 1–7 Full-Adders:} $S_i = A_i \oplus B_i \oplus C_i$, $C_{i+1} = (A_i \land B_i) \lor (B_i \land C_i) \lor (A_i \land C_i)$ & 
\textbf{Half-Adder (LSB)} \newline
Verilog:
\begin{verbatim}
module half_adder(input a, b, o
utput sum, carry);
  assign sum = a ^ b;
  assign carry = a & b;
endmodule
\end{verbatim}
VHDL:
\begin{verbatim}
entity half_adder is
  port(a, b: in std_logic; 
  sum, carry: out std_logic);
end entity;
architecture rtl of half_adder is
begin
  sum <= a xor b;
  carry <= a and b;
end architecture;
\end{verbatim}
\textbf{Full-Adder (bits 1–7)} \newline
Verilog:
\begin{verbatim}
module full_adder(input a, b, cin, 
output sum, cout);
  assign sum = a ^ b ^ cin;
  assign cout = (a & b) | (b & cin) | (a & cin);
endmodule
\end{verbatim}
VHDL:
\begin{verbatim}
entity full_adder is
  port(a, b, cin: in std_logic; sum, 
  cout: out std_logic);
end entity;
architecture rtl of full_adder is
begin
  sum <= a xor b xor cin;
  cout <= (a and b) or (b and cin) or (a and cin);
end architecture;
\end{verbatim} \\
\bottomrule
\end{tabular}
\caption{Qualitative example for 8-bit ripple carry adder. Symbolic decomposition shows Boolean/logical expressions for each submodule. Retrieved context contains modular Verilog and VHDL code snippets corresponding to these submodules. All submodules passed simulation/testbench.}
\label{tab:qualitative-rippleadder-symbolic}
\end{table*}

\section{Implementation Details \& Computation Cost}
\label{app:impl}
Our system is implemented in Python using the PyTorch framework, enabling flexible model development and efficient training. Retriever fine-tuning is performed on two NVIDIA V100 GPUs, which allows for effective processing our \textsc{RTL-IR} dataset. We orchestrate LLM queries using LangChain, and index high-dimensional vector representations with Milvus, a high-performance vector database offering both in-memory and GPU-accelerated similarity search. In contrast, traditional search solutions such as Elasticsearch \cite{elasticML2024} and Elastic's Learned Sparse Encoder Representations (ELSER) \cite{elasticML2024} serve as baselines; while Elasticsearch excels in full-text search, its semantic retrieval is limited compared to Milvus and ELSER. 

Our \textsc{SymDiReC} pipeline processes multiple prompts per task in parallel, achieving an average turnaround time of approximately 5--10 seconds. This efficiency demonstrates the practical viability of our approach for RTL code synthesis and summarization tasks. These implementation choices align with recent literature on retrieval-augmented generation \cite{lewis2020retrieval, guu2020retrieval} and domain-specific fine-tuning strategies. The integration of advanced indexing via Milvus and query orchestration using LangChain not only outperforms traditional retrieval methods but also substantially enhances the overall performance of our system.






\end{document}